\documentclass[letterpaper, 10 pt, conference]{ieeeconf}  % Comment this line out if you need a4paper
\usepackage{array}

\usepackage{amsmath,amssymb}
\usepackage{mathtools}
\usepackage{amsmath, amsfonts, amssymb} 
\usepackage{nicefrac}       % compact symbols for 1/2, etc.
\usepackage{microtype}      % microtypography
\usepackage{color,xcolor}
\usepackage{algorithm}
\usepackage{booktabs}
\usepackage{multirow} % add this in your preamble
\usepackage{enumerate}
\usepackage{algorithmic}
\usepackage{subfig}
\usepackage{graphicx}
\usepackage{balance}

\IEEEoverridecommandlockouts                              % This command is only needed if 
\title{\LARGE \bf
A Recipe for Efficient Sim-to-Real Transfer in Manipulation with Online Imitation–Pretrained World Models
}

\author{
Yilin Wang\textsuperscript{1\dag}, 
Shangzhe Li\textsuperscript{2\dag},
Haoyi Niu\textsuperscript{3},
Jiawei Fu\textsuperscript{1}, 
Zhiao Huang\textsuperscript{4}, 
Weitong Zhang\textsuperscript{2},
Hao Su\textsuperscript{4}%
\thanks{\dag\ Equal contributions.}%
\thanks{\textsuperscript{1}Yilin Wang and Jiawei Fu are with the Department of Computer Science and Engineering, University of California, San Diego.}%
\thanks{\textsuperscript{2}Shangzhe Li is with the Department of Computer Science, and 
Weitong Zhang is with the School of Data Science and Society, 
University of North Carolina at Chapel Hill.}%
\thanks{\textsuperscript{3}Haoyi Niu is with the Department of Mechanical Engineering, University of California, Berkeley.}
\thanks{\textsuperscript{4}Zhiao Huang and Hao Su are with Sudo AI GmbH.}%
}

\begin{document}

\maketitle
% \IEEEpeerreviewmaketitle
\thispagestyle{empty}
\pagestyle{empty}

%%%%%%%%%%%%%%%%%%%%%%%%%%%%%%%%%%%%%%%%%%%%%%%%%%%%%%%%%%%%%%%%%%%%%%%%%%%%%%%%
\begin{abstract}

% Sim-to-real transfer facilitates efficient policy learning by reducing the need for costly interactions with real-world environments. In this work, we study pure imitation learning in the sim-to-real setting with limited real data and no rewards, where existing offline imitation approaches often suffer from poor coverage and severe degradation. To overcome this challenge, we introduce a sim-to-real framework based on world models, which integrates online imitation pretraining with offline finetuning. By leveraging online interactions, our approach alleviates data coverage limitations, improving robustness, reducing performance degradation during finetuning, and enhancing generalization in domain transfer compared to offline methods. Empirically, it improves success rates by at least 31.7\% in sim-to-sim transfer and 23.3\% in sim-to-real transfer over existing offline imitation learning baselines.

We are interested in solving the problem of imitation learning with a limited amount of real-world expert data. Existing offline imitation methods often struggle with poor data coverage and severe performance degradation. We propose a solution that leverages robot simulators to achieve online imitation learning. Our sim-to-real framework is based on world models and combines online imitation pretraining with offline finetuning. By leveraging online interactions, our approach alleviates the data coverage limitations of offline methods, leading to improved robustness and reduced performance degradation during finetuning. It also enhances generalization during domain transfer. Our empirical results demonstrate its effectiveness, improving success rates by at least 84.3\% in sim-to-sim transfer and 113.3\% in sim-to-real transfer over existing offline imitation learning baselines.
\end{abstract}

%%%%%%%%%%%%%%%%%%%%%%%%%%%%%%%%%%%%%%%%%%%%%%%%%%%%%%%%%%%%%%%%%%%%%%%%%%%%%%%%
\section{INTRODUCTION}
Training robotic policies in simulation or from offline datasets and then finetuning them in the real world is a common strategy for real-world robotic control \cite{feng2023finetuning,do2025watch,ding2021sim,lin2025sim,cai2025navdp,maddukuri2025sim}. However, many of these works either require explicitly defined reward functions \cite{do2025watch,feng2023finetuning}, or depend on large-scale offline expert datasets \cite{cai2025navdp,maddukuri2025sim}. In contrast, we study a sim-to-real setting with limited expert demonstrations in both domains and no reward signals available. % The agent may only interact with the simulator without receiving rewards. 

This setting presents significant challenges for offline imitation learning % methods such as diffusion policies 
\cite{chi2023diffusionpolicy}, as limited expert coverage leads to overfitting and exacerbates bias accumulation; reinforcement learning approaches \cite{haarnoja2018soft,schulman2017proximal,hansen2024tdmpc2} are also inapplicable due to the absence of reward signals.

To address this, we propose a two-phase sim-to-real pipeline: (i) online imitation pretraining in simulation with simulated expert demonstrations and (ii) offline imitation finetuning with limited real-world demonstrations. During the pretraining phase, our method leverages a latent world model for efficient learning and employs the CDRED reward model \cite{li2025coupled} to generate reward signals by discriminating expert data from online interactions. Empirically, we show that online pretraining improves state-space coverage, reduces degradation after finetuning, and enhances out-of-distribution generalization. We further leverage a practical state estimation pipeline that fuses robot proprioceptive states with object pose estimation, enabling real-world deployment. We evaluate our approach on seven sim-to-sim transfer environments with two domain gaps and three sim-to-real transfer tasks. Our method achieves at least 84.3\% success rate improvement in sim-to-sim transfer and 113.3\% improvement in sim-to-real transfer over state-of-the-art offline imitation learning baselines.

Our main contributions are as follows:
(i) Proposing an efficient sim-to-real framework for limited-data regimes via online imitation pretraining of world models.
(ii) Demonstrating, through comparisons with existing offline/online imitation pretraining baselines, that our method achieves superior out-of-distribution (OOD) generalization and is more robust to performance degradation after few-shot finetuning.
(iii) Providing an empirical analysis that attributes these performance gains to the superior data coverage achieved during the online exploration phase.

% \hao{formulating your contributions in a more crisp manner.}

\section{RELATED WORKS}
\paragraph{Online Imitation Learning}
Online imitation learning is a variant of imitation learning that leverages reward-free interactions with the environment to improve performance, particularly when the size of the expert dataset is limited. Generative Adversarial Imitation Learning (GAIL) \cite{Ho2016GenerativeAI} introduces a framework based on generative adversarial training to perform online imitation learning. Subsequent works \cite{garg2021iqlearn,xiao2019wassersteinadversarialimitationlearning,alhafez2023} focus on stabilizing and improving GAIL within the adversarial imitation learning framework. Soft-Q Imitation Learning (SQIL) \cite{Reddy2019SQILIL} reformulates imitation learning as a reinforcement learning problem by assigning rewards of 1 to expert demonstrations and 0 to all other transitions. More recent approaches leverage model-based architectures to improve sample efficiency and performance, including standard model-based methods \cite{ren2024hybrid} and latent world model approaches \cite{efficientimitate,li2025reward,li2025coupled}.
\paragraph{Sim-to-Real Transfer for Manipulation}
Sim-to-real transfer aims to deploy a robotics policy trained in simulation to the real world by addressing domain gaps between the two environments \cite{niu2024comprehensive}. In robotic manipulation, prior works have explored a variety of approaches, including real-to-sim-to-real transfer with 3D reconstruction and point cloud–based policies \cite{torne2024reconciling}, sim-to-real transfer for vision-based humanoid dexterous manipulation via a divide-and-conquer policy distillation framework \cite{lin2025sim}, reinforcement learning–based transfer for deformable object manipulation \cite{pmlr-v87-matas18a}, residual learning for efficient contact-rich manipulation \cite{zhang2023efficient}, reinforcement learning methods for generalizable articulated object manipulation \cite{do2025watch}, vision-language model–generated rewards for real-to-sim-to-real transfer \cite{patel2025real}, and  diverse synthetic data generation for sim-to-real reinforcement learning \cite{wan2025lodestar}.

Unlike existing reinforcement learning-based approaches, our method relies purely on imitation learning, eliminating the need for reward signals in both simulation and the real world. Furthermore, in contrast to offline imitation learning methods that use demonstration datasets only, our approach leverages online interactions in a simulator to achieve broader data coverage through active exploration.

% \hao{add some sentences to contrast your method versus literature.}
\begin{figure*}[t]
\vspace{0.2cm}
    \centering
    \includegraphics[width=0.95\linewidth]{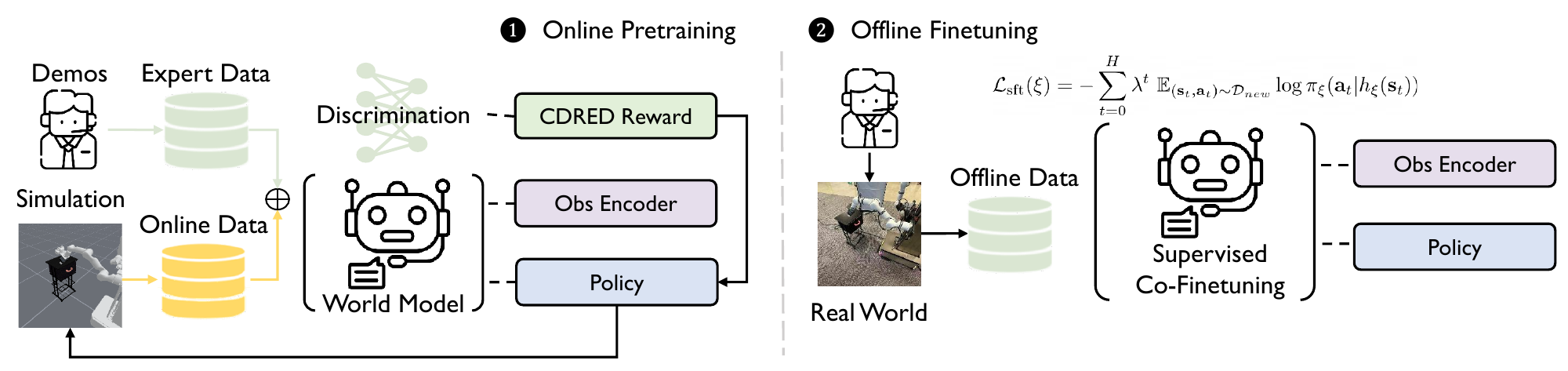}
    \vspace{-5pt}
    \caption{\textbf{Sim-to-Real Pipeline with Online Imitation Pretraining.} We illustrate the pretraining and finetuning pipeline of our proposed method. During pretraining, the world model is trained using a reward signal from a jointly trained discriminator that distinguishes expert from online interaction data. In the finetuning phase, the pretrained encoder and policy are refined using a small dataset of real-world expert demonstrations.}
    \label{fig:pipeline}
    \vspace{-15pt}
\end{figure*}

\section{PRELIMINARY}
\subsection{Problem Formulation}
We formulate our problem as a Markov Decision Process, defined by the tuple $\langle\mathcal{S}, \mathcal{A}, \mathcal{P}, r, \gamma\rangle$, where $\mathcal{S}$ is the state space; $\mathcal{A}$ is the action space; $\mathcal{P} : \mathcal{S} \times \mathcal{A} \rightarrow \Delta_{\mathcal{S}}$ is the transition function; $r : \mathcal{S} \times \mathcal{A} \rightarrow \{0,1\}$ is a sparse reward function that encodes binary success or failure feedback—assigning a reward of $1$ to the final state-action pair of a successful trajectory, and $0$ otherwise.$\gamma \in [0, 1]$ is the discount factor. Our goal is to learn a policy $\pi : \mathcal{S} \rightarrow \Delta_{\mathcal{A}}$ that minimizes the statistical distance between the state-action distribution induced by the policy, $\rho_\pi$, and the expert distribution $\rho_E$, using a set of expert state-action transitions in the simulator. We then aim to transfer the learned policy $\pi$ to the real world, where both the transition and reward function (unknown to the agent) may differ from those in the simulator.

\subsection{World Models}
In our setting, \emph{world model} is referring to reinforcement learning or imitation learning methods that explicitly learn a data-driven approximation of the environment’s transition dynamics—commonly known as model-based approaches. A key line of recent progress in this area is the development of latent-space world models \cite{hafner2019learning,hafner2019dream,hansen2022temporal,hansen2024tdmpc2}, which employ an encoder $\mathbf{z}=h(\mathbf{s})$ to map observations into a compact latent representation and a dynamics model $\mathbf{z}'=d(\mathbf{z},\mathbf{a})$ to predict transitions within that latent space. By predicting future latent states and generating imagined rollouts, these models enable efficient planning, leading to high sample efficiency and strong performance across a broad spectrum of challenging tasks \cite{hansen2024tdmpc2}.

% \begin{figure}
%     \centering
%     \includegraphics[width=0.95\linewidth]{real-pipeline.pdf}
%     \caption{\textbf{State Pre-processing during Real Deployment} We illustrate the state pre-processing pipeline used during real-world deployment. The object pose is estimated from RGBD camera observations using SAM2 and FoundationPose. This estimate is then combined with the robot’s proprioceptive states to form the input to the world model, which in turn generates actions for robotic control.}
%     \label{fig:state-preprocess}
% \end{figure}

\section{METHODOLOGY}
In this section, we introduce our sim-to-real framework designed for settings with \textbf{no rewards} and only \textbf{limited expert data} in both simulation and the real world. 
% The pipeline consists of two stages: online pretraining and offline finetuning. In pretraining, the agent learns from a small set of simulated expert trajectories and additional reward-free interactions in the simulator. In finetuning, reward signals remain absent; the agent is provided only with a limited set of successful real-world trajectories without rewards.

\subsection{Overview on the Setting}
In prior works, it is commonly assumed that a reward function is available in both simulation and the real world for reinforcement learning training \cite{do2025watch}, or that a sufficiently large offline expert dataset is accessible for imitation learning \cite{cai2025navdp}. In contrast, we consider a more challenging setting where no reward function is defined in both simulation and the real world. Instead, we rely on two small reward-free datasets, one from the simulation and one from the real environment, each consisting of expert demonstrations used for IL pretraining and finetuning, respectively. While interactions are permitted in the simulation environment, real-world finetuning depends solely on the limited expert dataset collected in the real world.

\subsection{Overview on the Method}
\begin{figure}
\vspace{0.2cm}
    \centering
    \includegraphics[width=0.95\linewidth]{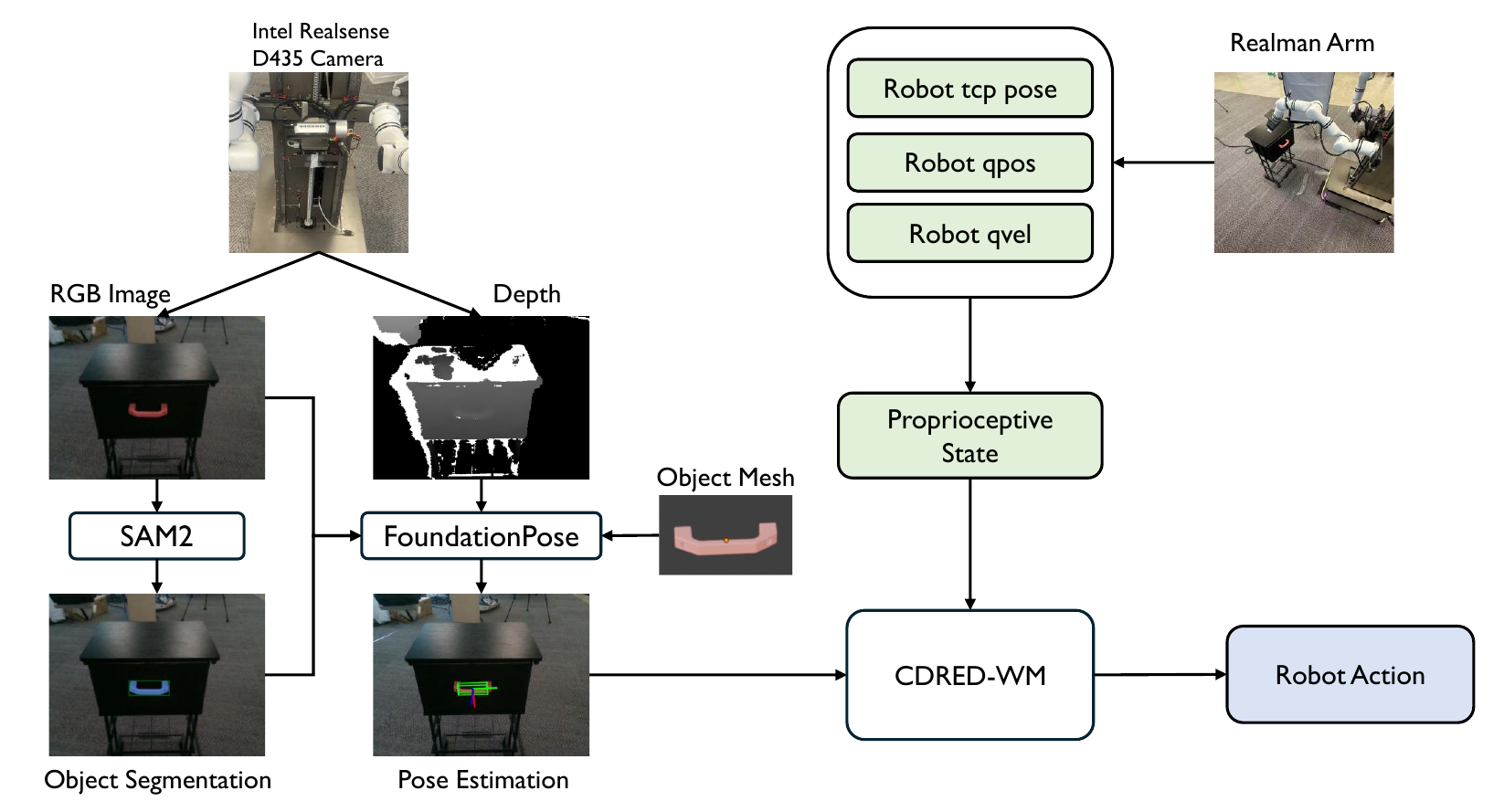}
    \vspace{-5pt}
    \caption{\textbf{State Pre-processing during Real Deployment.} We illustrate the state pre-processing pipeline used during real-world deployment. The object pose is estimated from RGBD camera observations using SAM2 and FoundationPose. This estimate is then combined with the robot’s proprioceptive states to form the input to the world model, which in turn generates actions for robotic control.}
    \vspace{-15pt}
    \label{fig:state-preprocess}
\end{figure}

The pipeline of our sim-to-real approach with a world model consists of three stages: online imitation pretraining in simulation, offline finetuning with real-world data, and real-world deployment. First, we pretrain the world model using CDRED reward model \cite{li2025coupled} to discriminate between expert data and online interactions in the simulated environment (Sec.~\ref{sec:pretrain}), leveraging state-based observations and 1M reward-free interactions. Next, we finetune the learned policy and encoder networks with a small dataset of expert demonstrations from the real world using a supervised finetuning objective (Sec.~\ref{sec:finetune}). The full pretraining and finetuning pipeline is illustrated in Fig.~\ref{fig:pipeline}. Finally, by using a state pre-processing pipeline that converts camera RGBD observations to real-time pose estimates, combined with proprioceptive feedback from the robot, we deploy our world model for real-world robotic control tasks (Sec.~\ref{sec:deploy}). The pipeline for state pre-processing in the real world is shown in Fig.~\ref{fig:state-preprocess}.

\subsection{World Model Online Imitation Pretraining}
\label{sec:pretrain}

Since no explicit reward signals are available during pretraining, standard RL approaches for world model training, such as TD-MPC2 \cite{hansen2024tdmpc2} and DreamerV3 \cite{hafner2023dreamerv3}, cannot be directly applied. To overcome this limitation, we adopt an \emph{online imitation learning} framework \cite{li2025reward,li2025coupled,yin2022planning}. In particular, we use the CDRED world model \cite{li2025coupled}, an imitation-based variant of TD-MPC2 that incorporates a reward model to align expert and behavioral distributions through a coupled distributional random expert distillation module.

\paragraph{Model architecture}  
The CDRED world model consists of five core components. An encoder first maps the input state $\mathbf{s}_t$ to a latent representation, $\mathbf{z}_t = h(\mathbf{s}_t)$. This latent state is then used by a latent dynamics model to predict the next state, $\mathbf{z}'_t = d(\mathbf{z}_t, \mathbf{a}_t)$, and a CDRED reward model to predict the immediate reward, $\hat{r}_t = R(\mathbf{z}_t,\mathbf{a}_t)$. For decision-making, the model also learns a value function, $\hat{q}_t = Q(\mathbf{z}_t, \mathbf{a}_t)$, and a policy, $\hat{\mathbf{a}}_t = \pi(\mathbf{z}_t)$, both of which operate on this latent state.

\paragraph{Reward model}  
The CDRED reward model consists of three parts: (i) an expert predictor $f_\phi$, (ii) a behavioral predictor $f_\psi$, and (iii) an ensemble of $K$ frozen target networks $\{f_{\bar\theta_k}\}_{k=0}^{K-1}$.The predictor and target networks estimate the densities of the expert and behavioral state-action distributions. The estimates are then used for reward construction. Following CDRED \cite{li2025coupled}, the reward is defined as:
% \vspace{-5pt}
\begin{equation}
\label{eqn:reward-formulation-modified}
R(\mathbf{z}_t,\mathbf{a}_t) = -\zeta~b(\mathbf{z}_t,\mathbf{a}_t, f_\phi) + (1 - \zeta)~b(\mathbf{z}_t,\mathbf{a}_t, f_\psi)
\end{equation}
where $\zeta$ is a tunable parameter balancing the expert distribution bonus (encouraging exploitation of expert data) and the behavioral distribution bonus (encouraging exploration).$b(\cdot)$ denotes the bonus term, which approximates the negative log-likelihood with respect to a distribution (behavioral or expert).  For example, if a datapoint $(\mathbf{z}, \mathbf{a})$ lies within the distribution represented by $f$, then $b(\mathbf{z}, \mathbf{a}, f)$ will be small.
% \niu{b should be better clarified more about what it is and how it works, while training details can refer to [9]. fixed.}

\paragraph{Training objective}  
Given an expert buffer $\mathcal{B}_E$ containing expert dataset $\mathcal{D}$ and a behavioral buffer $\mathcal{B}_\pi$ containing reward-free online rollouts, the world model is trained by minimizing:
% \vspace{-10pt}
\begin{align}
\label{eqn:model-loss}
\mathcal{L}(\phi,\psi,\xi) = \sum_{t=0}^H ~\mathbb{E}_{(\mathbf{s}_t,\mathbf{a}_t,\mathbf{s}'_t)\sim\mathcal{B}_E\cup\mathcal{B}_\pi}\Big[\mathcal{L}^{\text{base}} + \mathcal{L}^{\text{reward}}\Big],
\end{align}
with the overall objective comprises two parts: a reward model loss $\mathcal{L}^{\text{reward}}(\phi,\psi)$ and a joint loss $\mathcal{L}^{\text{base}}(\xi)$ for training the encoder, dynamics, and Q-function:
\begin{align*}
&\mathcal{L}^{\text{base}}(\xi) = \lambda^t\Big(\Vert\mathbf{z}'_t - \text{sg}(h_\xi(\mathbf{s}'_t))\Vert^2_2 + \text{CE}(\hat q_t, q_t)\Big), \\
&\mathcal{L}^{\text{reward}}(\phi,\psi) = \lambda^t\mathbb{E}_{k\sim\text{Uniform}(0,K)}\big(\text{T1}+\text{T2}\big),
\end{align*}
where $\text{CE}$ denotes the cross-entropy loss, following \cite{hansen2024tdmpc2,li2025coupled}, for stabilizing critic learning, and $\text{sg}$ denotes the stop-gradient operator. Here, the temporal difference target is computed with the CDRED reward:  
$q_t = R(\mathbf{z}_t,\mathbf{a}_t) + \gamma Q(\mathbf{z}'_t,\pi(\mathbf{z}'_t))$. The reward model objective is formulated as two MSE terms, analogous to the training objective of random network distillation \cite{burda2018exploration}, which estimate the densities of the expert and behavioral distributions for constructing the reward model:
\begin{align*}
\text{T1} &= \mathbb{E}_{(\mathbf{s}_t,\mathbf{a}_t)\sim\mathcal{B}_E}\Big[\Vert f_\phi(\mathbf{z}_t,\mathbf{a}_t) - f_{\bar\theta_k}(\mathbf{z}_t,\mathbf{a}_t)\Vert_2^2\Big], \\
\text{T2} &= \mathbb{E}_{(\mathbf{s}_t,\mathbf{a}_t)\sim\mathcal{B}_\pi}\Big[\Vert f_\psi(\mathbf{z}_t,\mathbf{a}_t) - f_{\bar\theta_k}(\mathbf{z}_t,\mathbf{a}_t)\Vert_2^2\Big].
\end{align*}

\paragraph{Policy optimization}  
A policy is then trained using the maximum-entropy objective similar to the one used in Soft Actor-Critic \cite{haarnoja2018soft} but with a discounted summation over the horizon $H$:
% \vspace{-5pt}
\begin{align}
\label{eqn:policy-learning}
% &\mathcal{L}^{\pi}(\xi)=\notag\\
% &\sum_{t=0}^H \lambda^t\Big[\mathbb{E}_{(\mathbf{s}_t,\mathbf{a}_t)\sim\mathcal{B}_E\cup\mathcal{B}_\pi}\big[-Q(\mathbf{z}_t,\pi_\xi(\mathbf{z}_t))+\beta\log\pi_\xi(\cdot|\mathbf{z}_t)\big]\Big].
\mathcal{L}^{\pi}(\xi)= \sum_{t=0}^H \lambda^t\mathbb{E}\big[-Q(\mathbf{z}_t,\pi_\xi(\mathbf{z}_t))+\beta\log\pi_\xi(\cdot|\mathbf{z}_t)\big], \notag 
\end{align}
where the expectation is taken over all state-action pairs sampled from $\mathcal{B}_E$ and  $\mathcal{B}_\pi$. 
For training stability, we stop gradients from the policy learning objective from propagating into the encoder, following \cite{hansen2024tdmpc2}.

\subsection{Offline Finetuning for Domain Transfer}
\label{sec:finetune}
After pretraining, we obtain a world model that performs well in the source domain. 
To transfer it to a new simulation domain or a real environment, we use a small expert dataset 
$\mathcal{D}_{\text{new}}$ collected from the target environment for supervised finetuning. 
Specifically, we adapt both the encoder $h_\xi$ and policy $\pi_\xi$ by optimizing a maximum likelihood estimation objective over the horizon $H$:
% \vspace{-10pt}
\begin{equation}
    \mathcal{L}_{\text{sft}}(\xi)=-\sum_{t=0}^H\lambda^t~\mathbb{E}_{(\mathbf{s}_t,\mathbf{a}_t)\sim\mathcal{D}_{\text{new}}}\log\pi_\xi(\mathbf{a}_t|h_\xi(\mathbf{s}_t)).
\end{equation}
Note that this finetuning is conducted fully offline, without any direct interaction with the new domain or the real environment. 
Instead, it only requires a small set of successful trajectories, which can be obtained, for example, through human teleoperation in the real world.

\begin{figure*}[t]
    \centering
    % ----- First row -----
    \vspace{0.2cm}
    \begin{minipage}{0.95\textwidth}
        \centering
        \includegraphics[width=\linewidth]{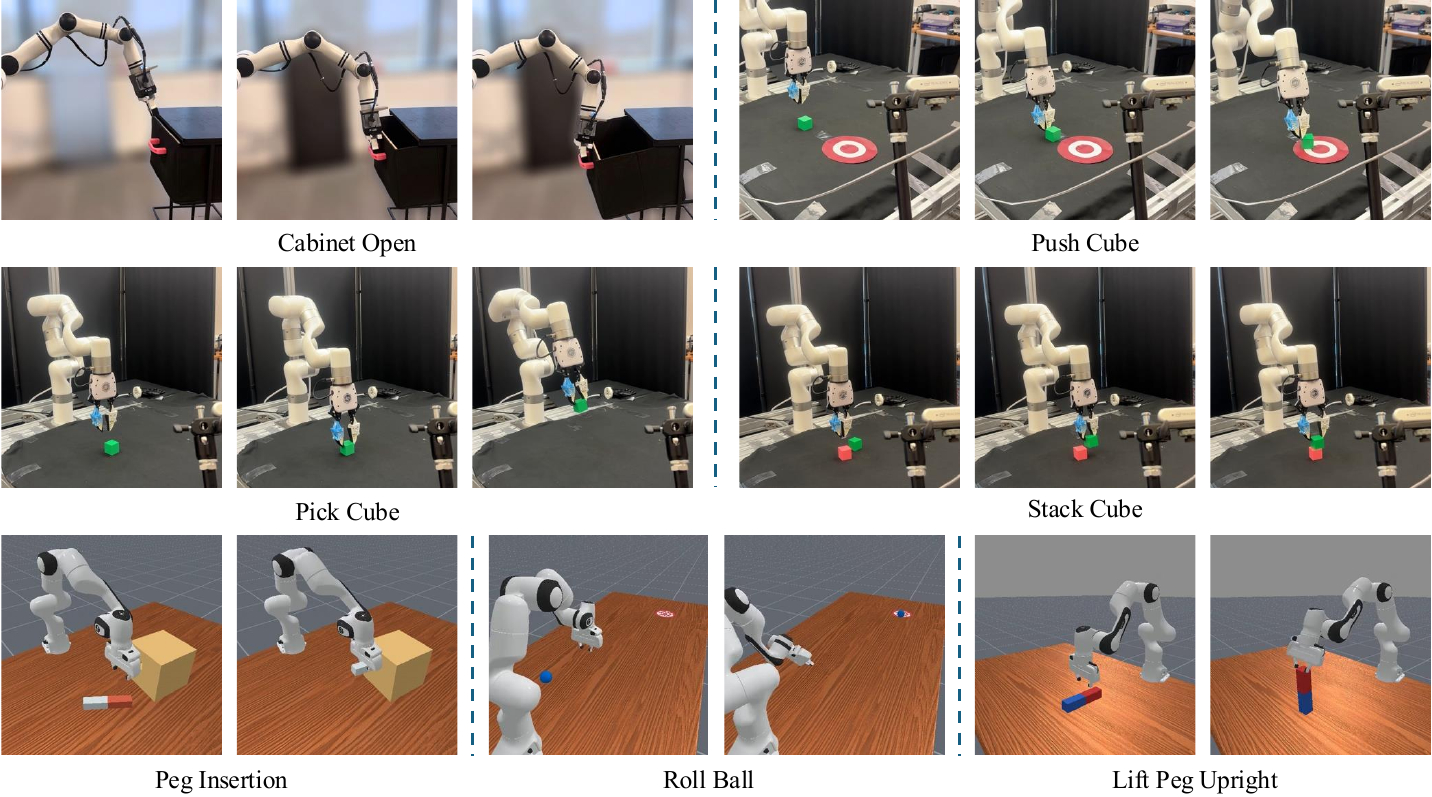}
    \end{minipage}
    \vspace{-3pt}
    \caption{\textbf{Illustrations of Tasks.} The first two rows show real-world demonstrations of the Cabinet Open, Push Cube, Pick Cube, and Stack Cube tasks, while the last row presents the initial and final successful states of the Peg Insertion, Roll Ball, and Lift Peg Upright tasks in simulation.}
    \label{fig:demos}
    \vspace{-10pt}
\end{figure*}

\subsection{State Estimation in the Real World}
\label{sec:deploy}

The states for our model consist of 2 parts: robot proprioceptive states and object (such as cube or handle) poses. For the robot proprioceptive states, we directly get the qpos (arm and gripper joint values) from the robot, which can then be used to compute the qvel (arm and gripper joint velocity) and the pose of tcp (the link that connects the gripper with the robot arm). 

For the object poses, an Intel RealSense D435 camera is used to estimate the object's pose. First, the camera captures the RGB image and the depth map that contain the target object. Later, the RGB image goes through a SAM2~\cite{ravi2024sam} model to get the object segmentation mask. The depth map, RGB image, segmentation mask as well as the object's 3D mesh are then fed into a FoundationPose~\cite{wen2024foundationpose} model to estimate and track the object poses. These poses are then combined with the robot proprioceptive states and fed into our CDRED-WM model to get the actions for the robot during real-world deployment. The pipeline is illustrated in Fig.~\ref{fig:state-preprocess}.

% \begin{figure}[htbp]
% \vspace{-10pt}
%   \centering
%   \subfloat[Customized Robot]{
%     \includegraphics[width=0.3\columnwidth]{figures/alpha4.jpg}
%     \label{fig:alpha4}
%   }
%   \subfloat[xArm7 Robot]{
%     \includegraphics[width=0.3\columnwidth]{figures/test_panda.png}
%     \label{fig:xarm7}
%   }
%   % \hfill % 这会在两张图之间插入水平空间
%   \subfloat[Franka Robot]{
%     \includegraphics[width=0.3\columnwidth]{figures/test_panda.png}
%     \label{fig:panda}
%   }
%   \vspace{-3pt}
%   \caption{\textbf{Illustration of Robots.} We show the three robot types used in our experiments: our customized robot, xArm7 and the Franka Panda. Since the Franka Panda arm is only used in simulation, we provide its simulated rendering, whereas for the other two robots we show real-world photo.}
%   \vspace{-15pt}
%   \label{fig:combined_figure}
% \end{figure}

\section{EXPERIMENTS}

\subsection{Robot Setup}
There are three robots involved in our tasks: a customized robot, a xArm7 robot and a Franka Emika Panda robot.

\begin{figure}[htbp]
\vspace{-15pt}
  \centering
  \subfloat[Customized Robot]{
    \includegraphics[width=0.3\columnwidth]{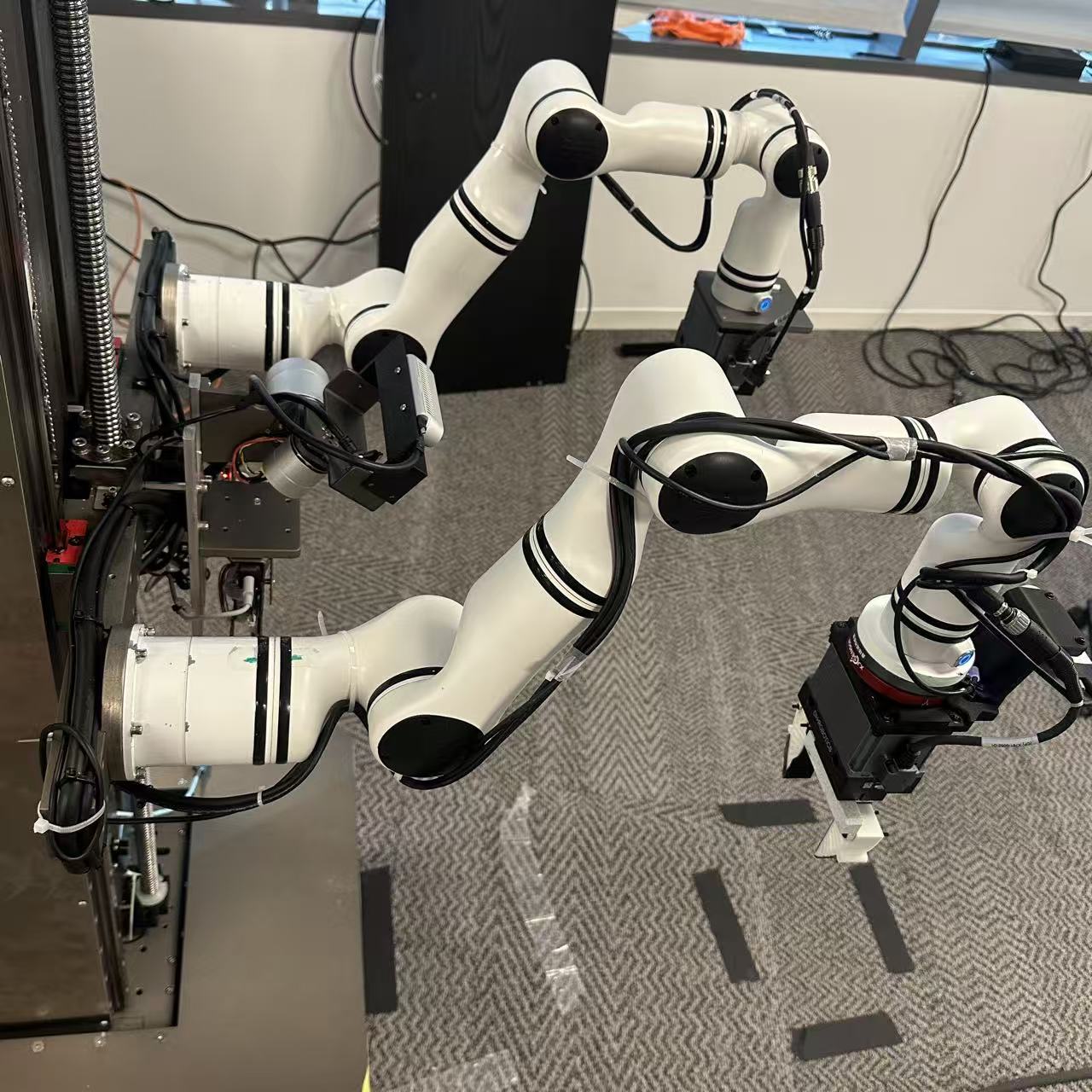}
    \label{fig:alpha4}
  }
  \subfloat[xArm7 Robot]{
    \includegraphics[width=0.3\columnwidth]{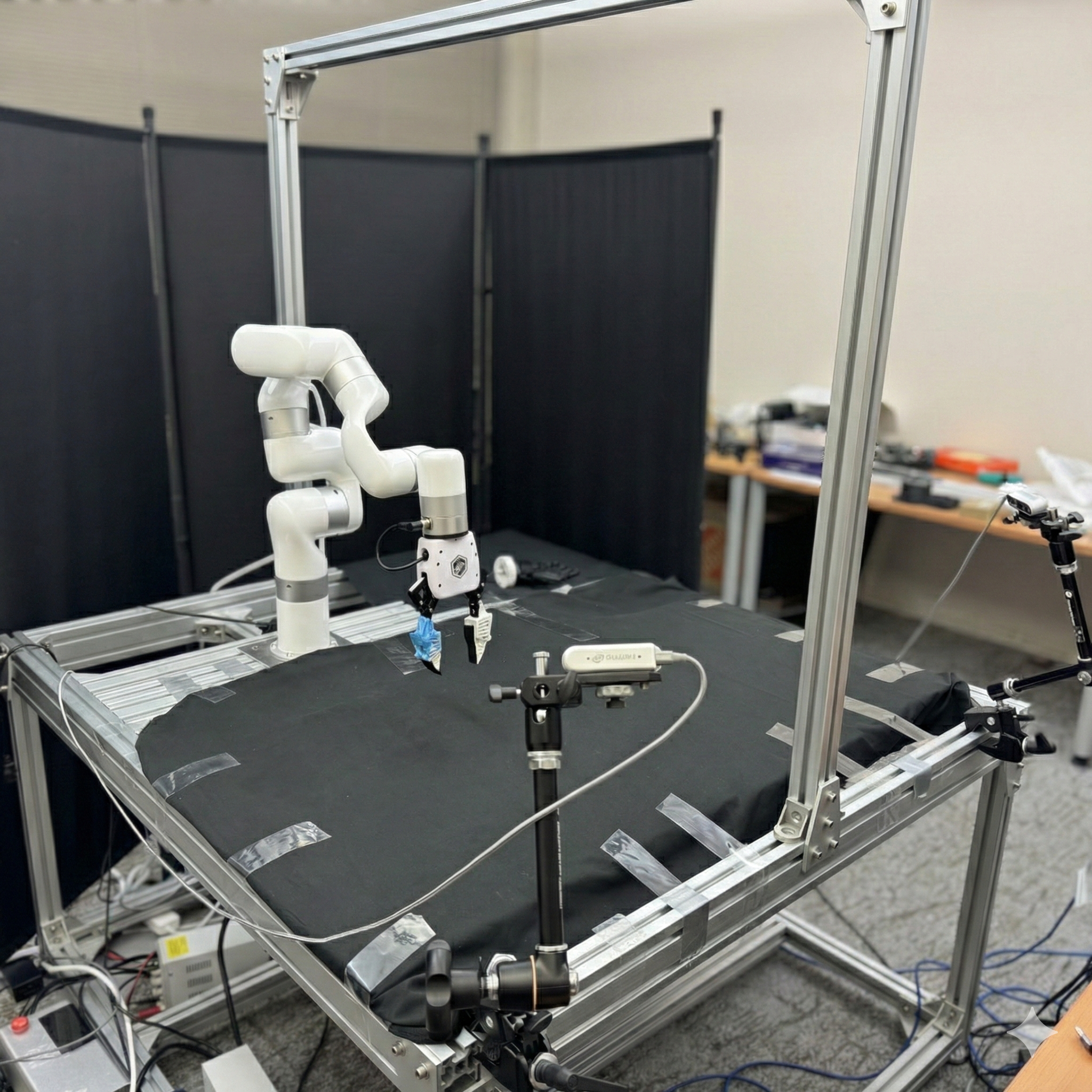}
    \label{fig:xarm7}
  }
  % \hfill % 这会在两张图之间插入水平空间
  \subfloat[Franka Robot]{
    \includegraphics[width=0.3\columnwidth]{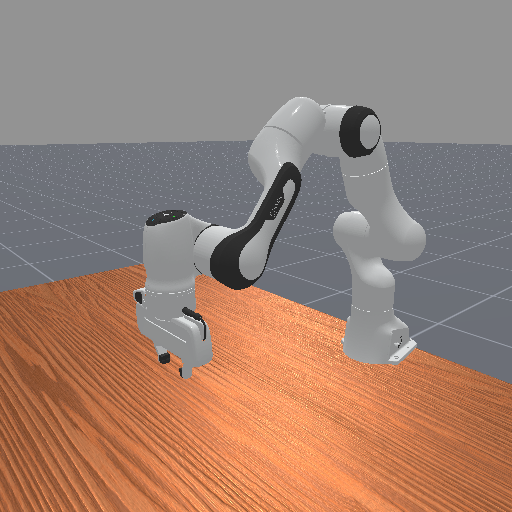}
    \label{fig:panda}
  }
  \vspace{-3pt}
  \caption{\textbf{Illustration of Robots.} We show the three robot types used in our experiments: our customized robot, xArm7 and the Franka Panda. Since the Franka Panda arm is only used in simulation, we provide its simulated rendering, whereas for the other two robots we show real-world photo.}
  \vspace{-20pt}
  \label{fig:combined_figure}
\end{figure}

Our customized robot features a height-adjustable rail, two 7-degree-of-freedom Realman arms (each with a 2-degree-of-freedom gripper), and an Intel RealSense D435 camera mounted on the rail (Fig.~\ref{fig:alpha4}). The rail joint value indicates the distance to the top in centimeters. In practice, we use the camera and one of the two arms mounted on the rail. The robot is controlled with PD control joint position scheme, and the action for arm is delta position, while the action for gripper is absolute position.
%The arm is controlled via PD joint delta pose control, and the gripper via PD joint position control.

The xArm7 robot consists of a 7-dof arm and a 6-dof gripper, which are controlled in the same way as our customized robot (Fig.~\ref{fig:xarm7}).

The Franka Emika Panda robot is a 7-dof arm with a 2-dof gripper. It is controlled in the same way as our customized robot. For our tasks, the Panda is fixed on a tabletop and used only in simulation (Fig.~\ref{fig:panda}).

\begin{table*}[t]
\vspace{0.2cm}
    \centering
    \begin{tabular}{c|c|c|cccc|ccc|c}
    \toprule
        Tasks & Robot & Domain Gap & BC (Direct) & BC (P+F) & DP (Direct) & DP (P+F) & SQIL & GAIL & IQ-MPC 
        & Ours \\
        \midrule
        \multirow{2}{*}{Cabinet Open} 
            & {Customized}& Biased Pose & 18\% & 47\% & 8\% & 9\% & 29\% & 6\% & 12\% & \textbf{62\%} \\
            &{Robot} & Rail Gap & 51\% & 51\% & 39\% & 36\% & 43\% & 14\% & 22\% & \textbf{66\%} \\
        \midrule
        {Push Cube} 
            & \multirow{3}{*}{xArm7} & \multirow{3}{*}{Biased Pose} & 65\% & 71\% & 14\% & 17\% & 48\% & 45\% & 58\% & \textbf{92\%} \\
        {Pick Cube} 
            & & & 15\% & 34\% & 23\% & 47\% & 24\% & 23\% & 26\% & \textbf{94\%} \\
        {Stack Cube} 
            & & & 2\% & 32\% & 34\% & 57\% & 27\% & 11\% & 28\% & \textbf{67\%} \\
        \midrule
        {Peg Insertion} 
            &\multirow{3}{*}{Franka} & \multirow{3}{*}{Biased Pose} & 8\% & 24\% & 0\% & 5\% & 33\% & 10\% & 51\% & \textbf{70\%} \\
        {Roll Ball} 
           & & & 0\% & 4\% & 32\% & 35\% & 21\% & 23\% & 37\% & \textbf{47\%} \\
        {Lift Peg Upright} 
           & & & 7\% & 58\% & 12\% & 25\% & 8\% & 16\% & 81\% & \textbf{93\%} \\ 
        \midrule
        Mean & - & - & 20.8\% & 40.1\% & 20.3\% & 28.9\% & 29.1\% & 18.5\% & 39.4\% & \textbf{73.9\%} \\
    \bottomrule
    \end{tabular}
    \vspace{-3pt}
    \caption{\textbf{Sim-to-Sim Transfer Results.} Results of sim-to-sim transfer experiments across multiple tasks under two domain gaps, biased pose and rail gap. The rail gap setting is specific to our customized robot platform, which includes an additional rail for controlling the vertical position of the arms. For offline imitation pretraining baselines, BC denotes behavior cloning with an encoder and Gaussian policy, and DP denotes diffusion policy. Direct indicates training only on the finetuning dataset $\mathcal{D}{\text{new}}$, while P+F denotes pretraining on the original domain expert dataset $\mathcal{D}$ followed by finetuning on $\mathcal{D}_{\text{new}}$. For online imitation learning baselines (SQIL, GAIL and IQ-MPC), all results use the P+F setting since online imitation learning cannot be performed using only offline $\mathcal{D}_{\text{new}}$ without environment interactions. }
    \label{tab:sim-to-sim-results}
    \vspace{-16pt}
\end{table*}

\subsection{Task Descriptions}
We evaluate the baselines and our approach on seven tasks in the sim-to-sim transfer experiments, four of which are also used in the sim-to-real transfer experiments. The details of these tasks are described below:
\begin{enumerate}[(i)]
    \item \textbf{Cabinet Open}. The task features a cabinet with a rail-mounted sliding drawer equipped with a front handle. Start from the closed position, the robot arm must reach the handle, secure it with the gripper, and pull the drawer open to about 50\% of its full rail length.The cabinet is randomly initialized within a $20 \times 20 \times 10\,\mathrm{cm^3}$ space (along the $x$, $y$, and $z$ axes) in simulation, and within a $15 \times 10\,\mathrm{cm^2}$ area (in the $x$--$y$ plane) in the real world. 
    %We purchased the cabinet on Amazon and constructed the 3D meshes correspond to it.
    \item \textbf{Push Cube}. The task involves manipulating a cube placed on the table surface using the robot arm. The arm must push the cube with its gripper until it reaches a specified target position on the table, which is designed as a circle with its center 20 cm away from the cube and a radius of 10 cm. The cube is randomly initialized within a $20 \times 20,\mathrm{cm^2}$ space (in the $x$--$y$ plane) in both simulation and real world.

    \item \textbf{Pick Cube}. The task requires the robot arm to grasp a cube placed on the tabletop and lift it to a specified target position, which is 15 cm above the cube. The cube is randomly initialized within a $20 \times 20,\mathrm{cm^2}$ space (in the $x$--$y$ plane) in simulation, and within a $15 \times 15\,\mathrm{cm^2}$ area (in the $x$--$y$ plane) in the real world.

    \item \textbf{Stack Cube}. The task requires the robot arm to grasp a cube placed on the tabletop, lift it up and place on another cube on the table. The cubes are randomly initialized within a $20 \times 20,\mathrm{cm^2}$ space (in the $x$--$y$ plane) in simulation, and within a $5 \times 10\,\mathrm{cm^2}$ area (in the $x$--$y$ plane) in the real world.

    \item \textbf{Peg Insertion}. The task requires the robot arm to grasp a peg with its gripper and insert it into a narrow hole of matching size. This task demands highly precise manipulation, and many offline imitation learning methods struggle with it due to bias accumulation. The peg is randomly initialized within a $20 \times 30\,\mathrm{cm^2}$ area (in the $x$--$y$ plane), and the box is initialized within a $10 \times 20\,\mathrm{cm^2}$ area in simulation. This task is evaluated only in simulation.

    \item \textbf{Roll Ball}. The task involves manipulating a ball placed on the table surface using the robot arm. The arm must push the ball with its gripper towards a randomly chosen target region on the table, defined as a circle located several decimeters away with a radius of 10\,cm. The ball is randomly initialized within a $30 \times 20\,\mathrm{cm^2}$ area (in the $x$--$y$ plane). This task is evaluated only in simulation.

    \item \textbf{Lift Peg Upright}. The task requires the robot arm to grasp a peg with its gripper and lift to any upright position on the table. The peg is randomly initialized within a $40 \times 40\,\mathrm{cm^2}$ area (in the $x$--$y$ plane) in simulation. This task is evaluated only in simulation.
    
\end{enumerate}
Cabinet Open task is performed with our customized robot, the three cube tasks are performed with xArm7 robot, whereas the other three use a Franka Panda arm. Illustrations of the tasks are shown in Fig.~\ref{fig:demos}.
% All of the tasks are performed using a Realman robot arm with 7 DoF. {\color{red} TODO: add a figure for illustration}

\subsection{Baselines}
For baselines, we evaluate two standard offline imitation learning methods: behavioral cloning (BC) with an encoder and a Gaussian policy, and diffusion policy (DP)~\cite{chi2023diffusionpolicy}. These baselines are included to highlight the effectiveness of our proposed online imitation pretraining approach.

Since our method involves online interaction with the original domain, we additionally compare against representative online imitation learning algorithms, including GAIL~\cite{Ho2016GenerativeAI}, SQIL~\cite{Reddy2019SQILIL}, and a strong model-based baseline~\cite{li2025reward}.

For offline IL baselines, we consider two training protocols:
\begin{enumerate}[(i)]
\item \textbf{Direct}: training solely on the finetuning expert dataset $\mathcal{D}{\text{new}}$ from the new domain.
\item \textbf{Pretrain + Finetune (P+F)}: pretraining on the expert dataset $\mathcal{D}$ from the original domain, followed by finetuning on $\mathcal{D}{\text{new}}$ from the new domain.
\end{enumerate}

For online IL methods, we only adopt the \textbf{Pretrain + Finetune (P+F)} setting, as direct training on the new domain is not applicable due to the lack of online interaction.

\subsection{Sim-to-Sim Transfer Results}
We investigate sim-to-sim domain adaptation in robot manipulation tasks within the ManiSkill3 simulation environments \cite{taomaniskill3}. For tasks involving the xArm7 and Franka robots, we model the domain gap through biased pose estimation: specifically, we introduce systematic perturbations to object poses in the target domain to simulate imperfect perception and pose estimation errors.

For Cabinet Open task that employs a customized robot with a linear rail, we introduce an additional domain shift. Beyond biased object pose estimation, we vary the robot’s rail position by shifting it to a new location (5cm below) relative to the pretrained environment. This rail displacement, combined with pose bias, creates a more substantial and structurally distinct domain gap compared to the source domain.  
% , considering two types of domain gaps:  
% \begin{enumerate}[(i)]
%     \item \textbf{Biased Pose Estimations}: a bias is introduced into the object poses of the new domain to simulate imperfect pose information.  
%     \item \textbf{Robot Rail Gap}: the robot arm’s rail is shifted to a new location along with biased pose estimations, distinct from that in the pretrained environment.  
% \end{enumerate}

\begin{table}[b]
\vspace{-10pt}
    \centering
    \begin{tabular}{c|c|c|c}
    \toprule
        Tasks & Pose Bias (cm) & $|\mathcal{D}|$ & $|\mathcal{D}_{new}|$\\
    \midrule
        Cabinet Open & (2, 2, 2) & 200 & 100 \\
        Push Cube & (2, 2, 1)  & 200 & 10 \\
        Pick Cube & (2, 2, 1)  & 200 & 30 \\
        Stack Cube & (2, 2, 1) & 400 & 50 \\
    \midrule
        Peg Insertion & (1, 1, 1) & 400 & 100 \\
        Roll Ball & (2, 2, 1) & 400 & 100 \\
        Lift Peg Upright & (2, 2, 2) & 100 & 10 \\
    \bottomrule
    \end{tabular}
    \caption{\textbf{Sim-to-Sim domain gaps and number of expert data used in sim-to-sim tasks.} $\mathcal{D}$ and $\mathcal{D}_{new}$ refers to origin and new domain dataset respectively. The 2 settings for Cabinet Open task use the same amount of data.}
    \label{tab:sim2sim-gap}
\end{table}

\begin{figure}[t]
\vspace{0.2cm}
    \centering
    \includegraphics[width=0.9\linewidth]{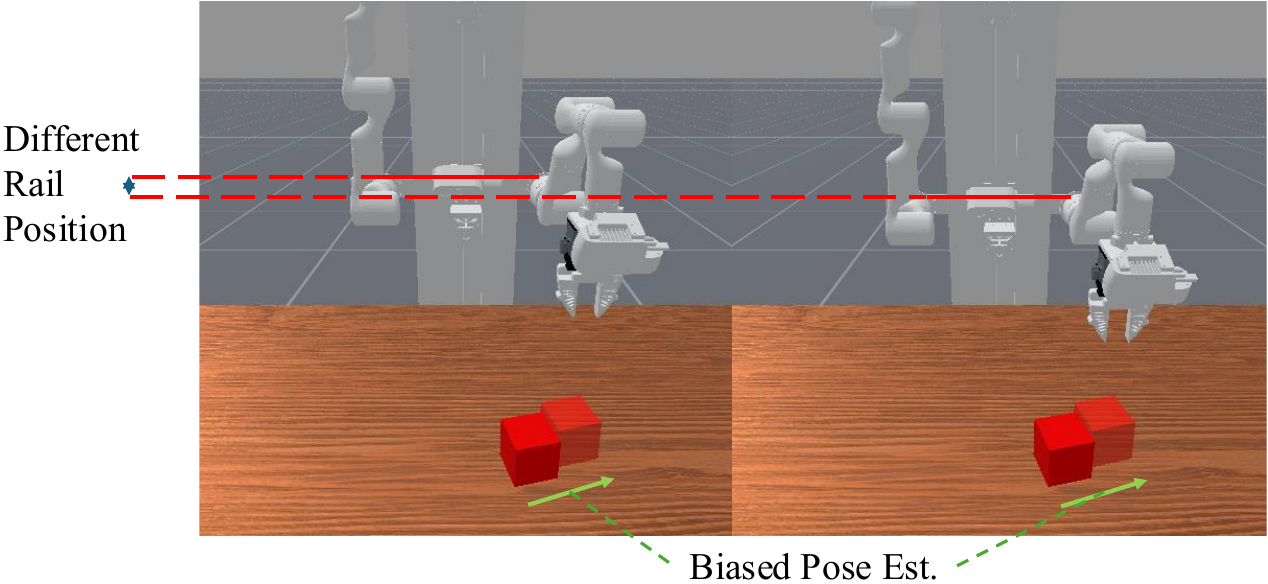}
    \vspace{-2pt}
    \caption{\textbf{Domain Gap in Sim-to-Sim Domain Transfer.} We illustrate the two types of domain gaps considered in our sim-to-sim transfer experiments. During finetuning in the target domain, the policy must adapt to either biased pose estimation or a shifted rail position (customized robot only).}
    \vspace{-20pt}
    \label{fig:domain-gap}
\end{figure}

Both types of gaps capture realistic challenges: pose estimation is often subject to noise, and the actual height of a robot on the rail may not perfectly align with its simulated counterpart. The gaps are illustrated in Fig.~\ref{fig:domain-gap} and Tab.~\ref{tab:sim2sim-gap}. Since the Franka Panda robot does not use a rail, we conduct experiments only with biased pose estimation for tasks involving this robot. The number of expert trajectories used for pretraining and finetuning is detailed in Tab.~\ref{tab:sim2sim-gap}. For expert data sampling in both pretraining and finetuning phases, we use a trained TD-MPC2~\cite{hansen2024tdmpc2} for the Pick Cube task, PPO~\cite{schulman2017proximal} for the Roll Ball and Stack Cube tasks, and SAC~\cite{haarnoja2018soft} for all remaining tasks.

We evaluate both the baselines and our method over 100 episodes, reporting the success rates in Tab.~\ref{tab:sim-to-sim-results}. Results show that in all of the 8 different sim-to-sim tasks, our CDRED-WM method can achieve the best performance when comparing with baseline methods with existing offline or online imitation pretraining baselines, while our method can achieve at least 10\% better results than the second best methods, among which Push Cube and Pick Cube settings experience more than 30\% performance boost. 
% {\color{red} TOOO: rewrite the score summary, also in the abstract} 
The comparison between our approach and existing offline/online imitation pretraining methods highlights the superior generalization capability of our proposed online imitation pretraining framework. A more detailed analysis is provided in Sec.~\ref{sec:online-imitation-pro}. Compared with other online imitation pretraining approaches, methods such as GAIL~\cite{Ho2016GenerativeAI} and IQ-MPC~\cite{li2025reward} rely on adversarial formulations, which are typically less stable than the density-estimation-based reward modeling adopted in our domain adaptation framework~\cite{li2025coupled}. Furthermore, adversarial training often induces sharper reward distributions due to the min--max interaction between the policy and the discriminator, which may negatively affect robustness during domain transfer. Moreover, SQIL \cite{Reddy2019SQILIL} provides only sparse learning signals, assigning binary rewards (0 or 1) to behavioral and expert data, respectively. Such sparse supervision makes pretraining particularly challenging for complex tasks. In contrast, our method instead leverages a learned reward model via density estimation, yielding smoother and more informative learning signals, which leads to improved stability during pretraining and stronger downstream performance. 
% \begin{table}[b]
% \vspace{-10pt}
%     \centering
%     \begin{tabular}{c|ccc}
%     \toprule
%         Dataset & $\mathcal{D}$ & \multicolumn{2}{c}{$\mathcal{D}_{new}$} \\
%     \midrule
%         Domain & Original & Biased Pose & Kinematics Gap \\
%     \midrule
%         Cabinet Open & 200  & 100 & 100 \\
%         Push Cube & 200 & 10 & - \\
%         Pick Cube & 200 & 30 & - \\
%         Stack Cube & 400 & 50 & - \\
%     \midrule
%         Peg Insertion & 400 & 100 & - \\
%         Roll Ball & 400 & 100 & - \\
%         Lift Peg Upright & 100 & 10 & - \\
%     \bottomrule
%     \end{tabular}
%     \caption{Number of expert data used in sim-to-sim tasks.}
%     \label{tab:sim-to-sim-experts}
% \end{table}

\begin{table*}[t]
\vspace{0.2cm}
    \centering
    \begin{tabular}{c|cccc|ccc|c}
    \toprule
        Tasks & BC (Direct) & BC (P+F) & DP (Direct) & DP (P+F) & SQIL & GAIL & IQ-MPC 
        & CDRED-WM (Ours) \\
        \midrule
        {Cabinet Open} 
             & 6/10 & 1/10 & 4/10 & 3/10 & 4/10 & 0/10 & 1/10 & \textbf{9/10} \\

        {Push Cube} 
             & 2/10 & 2/10 & 0/10 & 0/10 & 9/10 & 6/10 & 7/10 & \textbf{10/10} \\
        {Pick Cube} 
             & 5/10 & 6/10 & 0/10 & 1/10 & 0/10 & 0/10 & 2/10 & \textbf{8/10} \\
        {Stack Cube} 
             & 0/10 & 0/10 & 2/10 & 4/10 & 2/10 &  0/10 & 1/10 & \textbf{5/10} \\
    \midrule
    Total & 13/40 & 9/40 & 6/40 & 8/40 & 15/40 & 6/40 & 11/40 & \textbf{32/40} \\
    \bottomrule
    \end{tabular}
    \vspace{-3pt}
    \caption{\textbf{Sim-to-Real Transfer Results.} We evaluate our approach against baseline methods on three sim-to-real transfer tasks using our customized robot. Each task is assessed over 10 trajectories, and the success rate is reported. Our method achieves the highest success rate compared to existing offline imitation learning approaches.}
    \label{tab:sim-to-real-results}
    \vspace{-10pt}
\end{table*}

\subsection{Sim-to-Real Transfer Results}

We evaluate the following tasks in the sim-to-real setting: Cabinet Open, Push Cube, Pick Cube and Stack Cube. For each task, we use the simulation-pretrained CDRED world model as base model (same pretrained model as in sim-to-sim settings) and collect 15 (for Push Cube and Pick Cube) or 30 (for Cabinet Open and Stack Cube) real-world expert trajectories for finetuning. To obtain these trajectories, we first pretrain a CDRED world model in simulation and then incorporate human corrections into the direct rollouts of the pretrained model during real-world execution. We evaluate both the baselines and our method over 10 episodes, reporting the success rates in Tab.~\ref{tab:sim-to-real-results}. 

Results demonstrate that with same amount of real-world expert trajectories, our CDRED-WM method performs much better than the baseline methods on the Cabinet Open task. For the Push Cube and Pick Cube tasks, our CDRED method significantly outperforms the BC-based baselines, while slightly outperforms the DP-based baselines. Overall, CDRED-WM method can achieve more stable and better results when comparing with baseline methods on the three tasks, which shows its strong potential in sim-to-real.

\subsection{Additional Experimental Analysis}
\label{sec:online-imitation-pro}

In this section, we present additional experimental analyses of our sim-to-real framework from three perspectives: data coverage, performance degradation after finetuning, and out-of-distribution (OOD) generalization.

\begin{figure}[t]
\vspace{-10pt}
    \centering
    \begin{minipage}{0.32\linewidth}
        \centering
        \includegraphics[width=\linewidth]{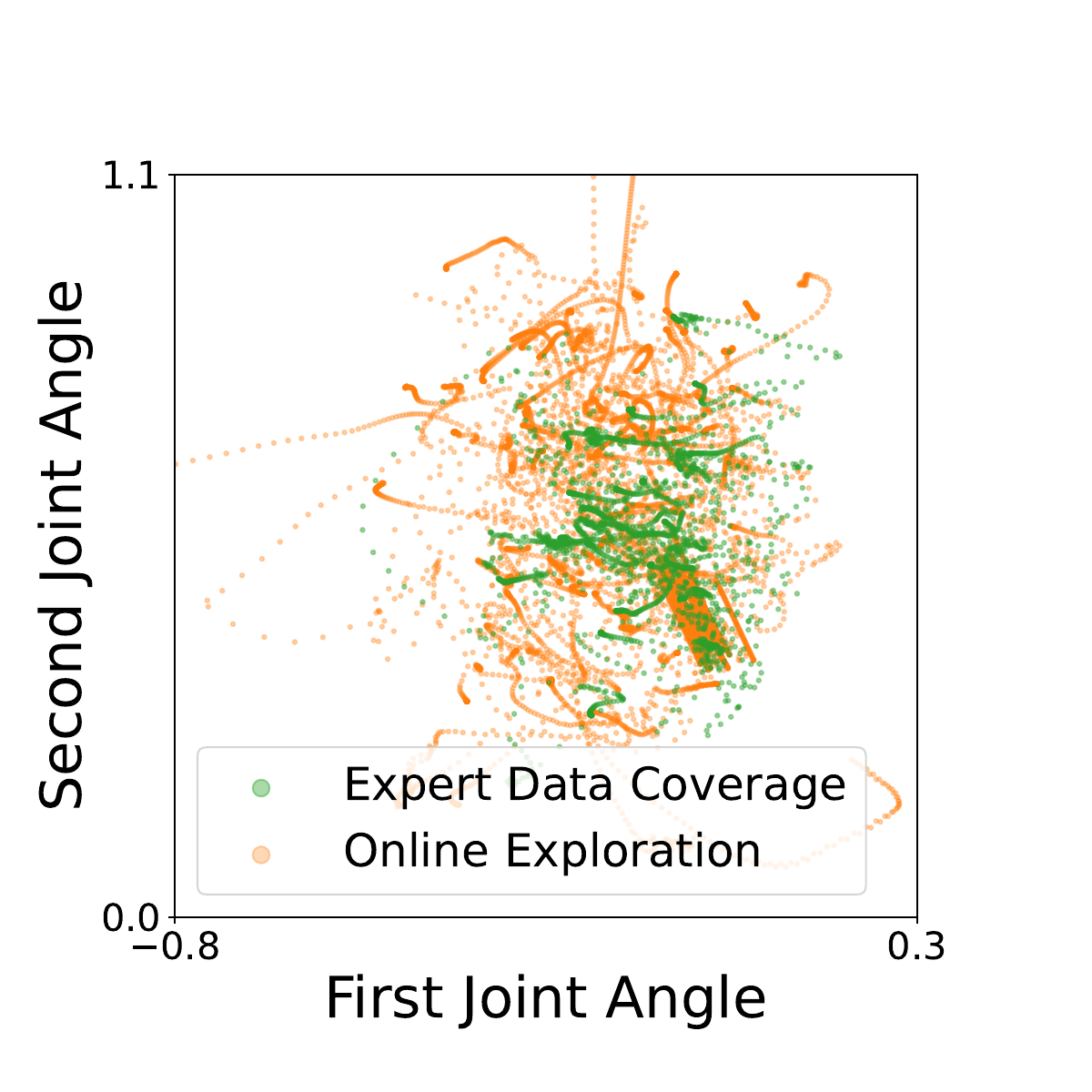}
    \end{minipage}
    \hfill
    \begin{minipage}{0.32\linewidth}
        \centering
        \includegraphics[width=\linewidth]{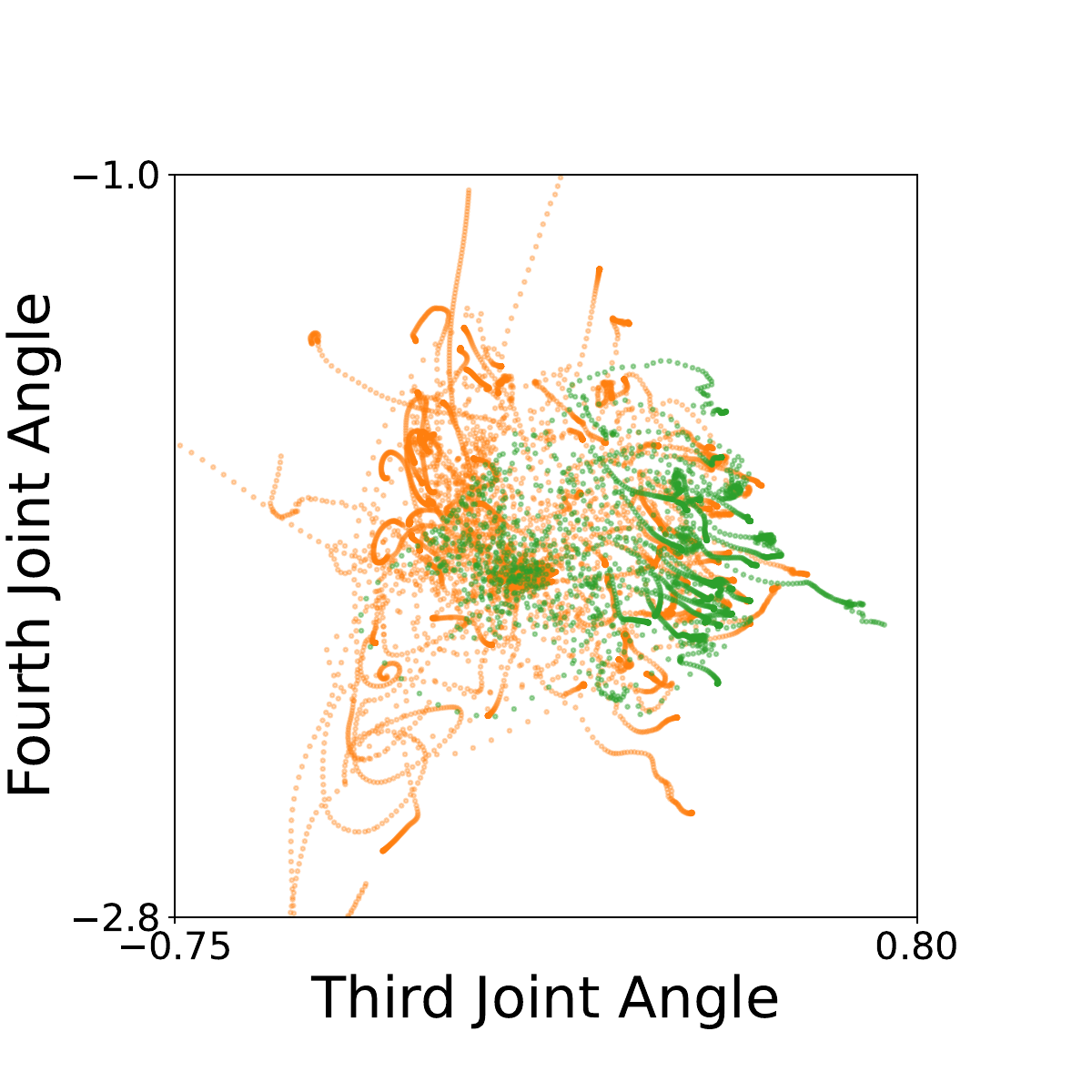}
    \end{minipage}
    \hfill
    \begin{minipage}{0.32\linewidth}
        \centering
        \includegraphics[width=\linewidth]{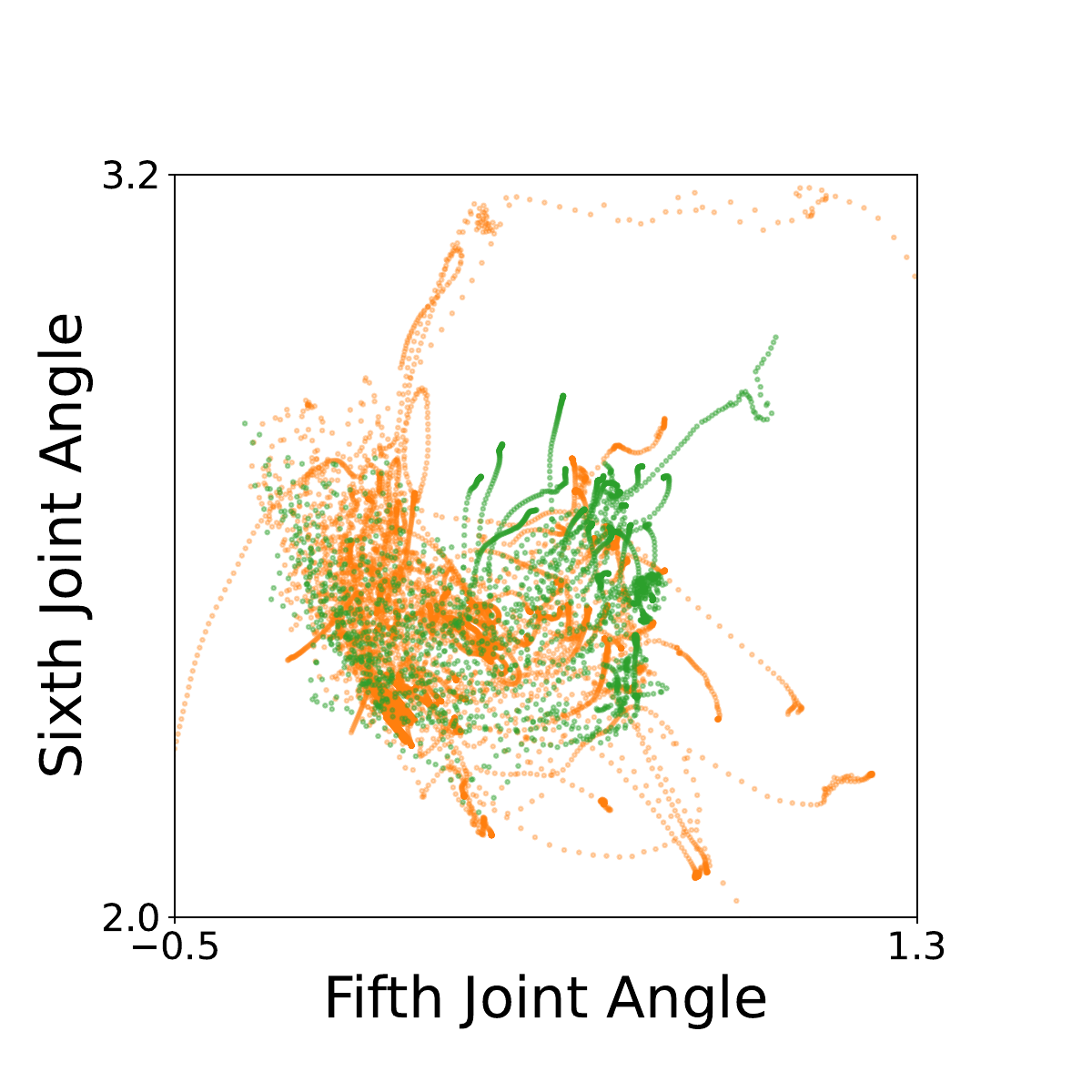}
    \end{minipage}
    \vspace{-7pt}
    \caption{\textbf{Data Coverage.} Visualization of state-space coverage from the expert dataset and online exploration using CDRED-WM in the Peg Insertion task, shown for six joint angles of the robot arm. The results indicate that online exploration effectively provides additional coverage in regions of the state space not represented in the expert dataset.}
    \vspace{-10pt}
    \label{fig:coverage}
\end{figure}

\paragraph{Better Coverage for Pretraining}
We first demonstrate the effectiveness of our sim-to-real pipeline by comparing it with offline imitation pretraining baselines, such as behavior cloning (BC)~\cite{bain1995framework} and diffusion policy~\cite{chi2023diffusionpolicy}, which directly fit the distribution of the offline expert dataset. When the expert dataset is small and lacks sufficient coverage, the resulting pretrained policy becomes narrowly distributed and may suffer from distributional collapse during finetuning. Incorporating online reward-free interactions during pretraining expands coverage of the state space, which improves generalization and mitigates bias accumulation (a key factor impacting the performance of offline IL in the Peg Insertion and Roll Ball tasks, as shown in Tab.~\ref{tab:sim-to-sim-results}). Fig.~\ref{fig:coverage} compares the coverage of the offline expert dataset with that of the online exploration data, illustrating how online interactions substantially broaden the explored state space.

% \begin{figure}[t]
% % \vspace{0.2cm}
%     \centering
%     \includegraphics[width=0.95\linewidth]{wucha.pdf}
%     \caption{\textbf{Domain Gap in Sim-to-Sim Domain Transfer.} We illustrate the two types of domain gaps considered in our sim-to-sim transfer experiments. During finetuning in the target domain, the policy must adapt to either biased pose estimation or a shifted rail position.}
%     \vspace{-10pt}
%     \label{fig:domain-gap}
% \end{figure}

\paragraph{Less Performance Degradation after Finetuning}
Compared to offline pretraining methods such as diffusion policy (DP) and behavior cloning (BC), as well as adversarial-based online imitation pretraining approaches GAIL~\cite{Ho2016GenerativeAI} and IQ-MPC~\cite{li2025reward}, and sparse-reward labeling method SQIL~\cite{Reddy2019SQILIL}, our online imitation learning approach exhibits less performance degradation under supervised finetuning with domain transfer. Results in Tab.~\ref{tab:degrade-finetune} show that existing offline/online IL pretraining baselines suffer significant drops in success rate after finetuning with limited expert data from the new domain, whereas our approach largely preserves performance.

A plausible explanation for our superiority over offline imitation learning methods is that online imitation pretraining encourages active exploration, resulting in broader coverage of the state–action space rather than narrowly fitting the distribution of the offline expert dataset. This broader coverage improves robustness to distribution shifts during domain transfer, especially when the expert pretraining dataset is relatively small. Prior work on offline-to-online reinforcement learning~\cite{zhou2024efficient,shin2025online} has similarly observed that directly finetuning policies pretrained purely offline can lead to performance degradation or distributional collapse when encountering out-of-distribution (OOD) transitions, largely due to limited policy support. These findings suggest that online pretraining provides inherent robustness advantages compared to purely offline imitation pretraining approaches.

When comparing our framework with existing online imitation pretraining methods, we attribute the advantage to our density-estimation-based reward modeling. Adversarial approaches such as GAIL~\cite{Ho2016GenerativeAI} and IQ-MPC~\cite{li2025reward} rely on discriminator-based formulations that may produce sharper pretrained policy distributions, which are more susceptible to performance degradation after adaptation. SQIL~\cite{Reddy2019SQILIL}, on the other hand, formulates imitation learning as sparse-reward reinforcement learning during pretraining. Although it avoids adversarial formulation and therefore exhibits relatively small degradation after finetuning, its sparse reward signal limits its ability to solve complex tasks, leading to weaker pretraining performance. In contrast, our method achieves both strong pretraining performance and improved stability during domain transfer, resulting in superior overall performance.
\begin{table}[h]
% \vspace{-10pt}
    \centering
    \begin{tabular}{c|cccccc}
    \toprule
         Domain & BC & DP & SQIL & GAIL & IQ-MPC & Ours \\
        \midrule
        Original & 82\% & 61\% & 50\% & 69\% & 92\% & 93\% \\
        New  & 71\% & 17\% & 48\% & 45\% & 58\% & 92\% \\
        \midrule
        Degradation & -11\% & -44\% & -2\% & -24\% & -34\% & -1\% \\
        \bottomrule
        
    \end{tabular}
    \vspace{-3pt}
    \caption{\textbf{Performance Degradation after Supervised Finetuning.} We analyze the performance degradation after finetuning on a new domain with incorrect state estimation gap compared to the performance on the original domain on Push Cube task. Results show that our online IL pretraining approach suffers less after finetuning on a new domain compared to existing offline/online IL pretraining approaches.}
    \label{tab:degrade-finetune}
    \vspace{-5pt}
\end{table}

\begin{table}[h]
% \vspace{-5pt}
    \centering
    \begin{tabular}{c|cccccc}
    \toprule
         Setting & BC & DP & SQIL & GAIL & IQ-MPC & Ours \\
        \midrule
        I.D. & 51\% & 36\% & 43\% & 14\% & 22\% & 66\% \\
        O.O.D. & 28\% & 8\% & 32\% & 0\% & 6\% & 62\% \\
        \midrule
        Degradation & -23\% & -28\% & -11\% & -14\% & -16\% & -4\% \\
        \bottomrule
        
    \end{tabular}
    \vspace{-3pt}
    \caption{\textbf{Generalization to Unseen Initializations.} We evaluate success rates over 100 trajectories by deploying the model on initializations not included in the finetuning dataset. The results show that our method exhibits less performance degradation, indicating stronger out-of-distribution generalization. “I.D.” denotes in-distribution settings, while “O.O.D.” denotes out-of-distribution settings.}
    \label{tab:unseen}
    \vspace{-10pt}
\end{table}

\paragraph{Stronger Out-of-distribution Generalization Ability}
Additionally, we evaluate the out-of-distribution generalization ability of our proposed domain adaptation approach. In the domain transfer phase, expert data from the new domain is collected from a narrow set of initial states, while evaluation is performed on unseen initializations. We study this setting using the Cabinet Open task, where the deployment initialization of the cabinet is shifted by 20cm, which is completely outside the range used for finetuning trajectory sampling. As shown in Tab.~\ref{tab:unseen}, our method achieves stronger performance on these unseen initializations compared to offline/online imitation pretraining baselines.

\subsection{Ablation Studies}
We conduct ablation study on the number of expert trajectories required during finetuning for our approach in simulation domain transfer. We conduct the evaluation on Cabinet Open task. Results in Tab.~\ref{tab:ablation-traj} show that our approach can still achieve 50\% success rate with only 20 expert trajectories from the new domain with biased state estimation.

\begin{table}[h]
\vspace{-5pt}
    \centering
    % \begin{tabular}{c|c}
    % \toprule
    %     Trajectory Number & Success Rate \\
    % \midrule
    %     100 & 62\% \\
    %     50 & 56\% \\
    %     20 & 50\% \\
    %     10 & 42\% \\
    %     \bottomrule
    % \end{tabular}
    \begin{tabular}{c|cccc}
    \toprule
        Number of Trajectories & 100 & 50 & 20 & 10 \\
    \midrule
        Success Rate & 62\% & 56\% & 50\% & 42\% \\
        \bottomrule
    \end{tabular}
    \vspace{-3pt}
    % \caption{\textbf{Ablation Study on Finetuning Trajectory Number.} We perform an ablation over the number of finetuning trajectories (100, 50, 20, and 10) and report success rates on 100 evaluation trajectories.}
    \caption{\textbf{Ablation Study.} We perform an ablation over the number of finetuning trajectories (100, 50, 20, and 10) and report success rates on 100 evaluation trajectories.}
    \label{tab:ablation-traj}
    \vspace{-15pt}
\end{table}

% \subsection{Ablation Studies}
% We conduct ablation study on the number of expert trajectories required during finetuning for our approach in simulation domain transfer. We conduct the evaluation on Cabinet Open task. Results in Tab.~\ref{tab:ablation-traj} show that our approach can still achieve 50\% success rate with only 20 expert trajectories from the new domain with biased state estimation.

\subsection{Hyperparameter Settings}

During finetuning, we optimize the encoder and policy using the supervised finetuning objective with a learning rate of $1\times10^{-5}$ and a weight decay of $1\times10^{-4}$. We use a batch size of 256 and a horizon of 3. Pretraining hyperparameters are the same as in the original CDRED paper \cite{li2025coupled}.

\section{CONCLUSIONS}
In this paper, we propose a sim-to-real framework based on world models for settings with no reward signals and only limited expert data in both simulation and the real world. Our approach employs a CDRED reward model to generate surrogate rewards by discriminating expert from behavioral data, thereby facilitating world model training. The pretrained model is then finetuned on a few-shot real-world expert dataset. We evaluate our method on seven sim-to-sim manipulation transfer tasks with two manually designed domain gaps, as well as three sim-to-real manipulation tasks on real robots. Experimental results demonstrate that our approach improves success rates by at least 84.3\% in sim-to-sim transfer and 113.3\% in sim-to-real transfer compared to state-of-the-art offline imitation learning baselines.

\addtolength{\textheight}{-12cm}   % This command serves to balance the column lengths

%%%%%%%%%%%%%%%%%%%%%%%%%%%%%%%%%%%%%%%%%%%%%%%%%%%%%%%%%%%%%%%%%%%%%%%%%%%%%%%%

%%%%%%%%%%%%%%%%%%%%%%%%%%%%%%%%%%%%%%%%%%%%%%%%%%%%%%%%%%%%%%%%%%%%%%%%%%%%%%%%



% \section*{ACKNOWLEDGMENT}

%%%%%%%%%%%%%%%%%%%%%%%%%%%%%%%%%%%%%%%%%%%%%%%%%%%%%%%%%%%%%%%%%%%%%%%%%%%%%%%%

% References are important to the reader; therefore, each citation must be complete and correct. If at all possible, references should be commonly available publications.

% \sloppy

\bibliographystyle{IEEEtran}
\bibliography{IEEEexample}

\end{document}